\documentclass[journal]{IEEEtran}

\ifCLASSINFOpdf
\else
   \usepackage[dvips]{graphicx}
\fi
\usepackage{url}

\usepackage{graphicx}
\usepackage{amsfonts,amsmath}
\usepackage[pagebackref=false,breaklinks=true,colorlinks,citecolor=black,linkcolor=black,bookmarks=false]{hyperref} 
\usepackage{float}
\usepackage{graphicx}
\usepackage{subfig}
\usepackage{amssymb}
\usepackage{multirow}
\usepackage{booktabs}

\begin{document}

\title{Hash Grid Feature Pruning}

\author{Yangzhi Ma, Bojun Liu, Jie Li, Li Li, and Dong Liu
\thanks{Y.~Ma, B.~Liu, J.~Li, L.~Li, and D.~Liu are with University of Science and Technology of China.}}

\maketitle

\begin{abstract}

Hash grids are widely used to learn an implicit neural field for Gaussian splatting, serving either as part of the entropy model or for inter‑frame prediction. However, due to the irregular and non‑uniform distribution of Gaussian splats in 3D space, numerous sparse regions exist, rendering many features in the hash grid invalid. This leads to redundant storage and transmission overhead. In this work, we propose a hash grid feature pruning method that identifies and prunes invalid features based on the coordinates of the input Gaussian splats, so that only the valid features are encoded. This approach reduces the storage size of the hash grid without compromising model performance, leading to improved rate–distortion performance. Following the Common Test Conditions (CTC) defined by the standardization committee, our method achieves an average bitrate reduction of 8\% compared to the baseline approach.
\end{abstract}

\begin{IEEEkeywords}
Gaussian Splatting Compression, Hash Grid, Feature Pruning. 
\end{IEEEkeywords}

\IEEEpeerreviewmaketitle

\section{Introduction}

\begin{figure}[t]
\centering
\includegraphics[width=0.99\linewidth]{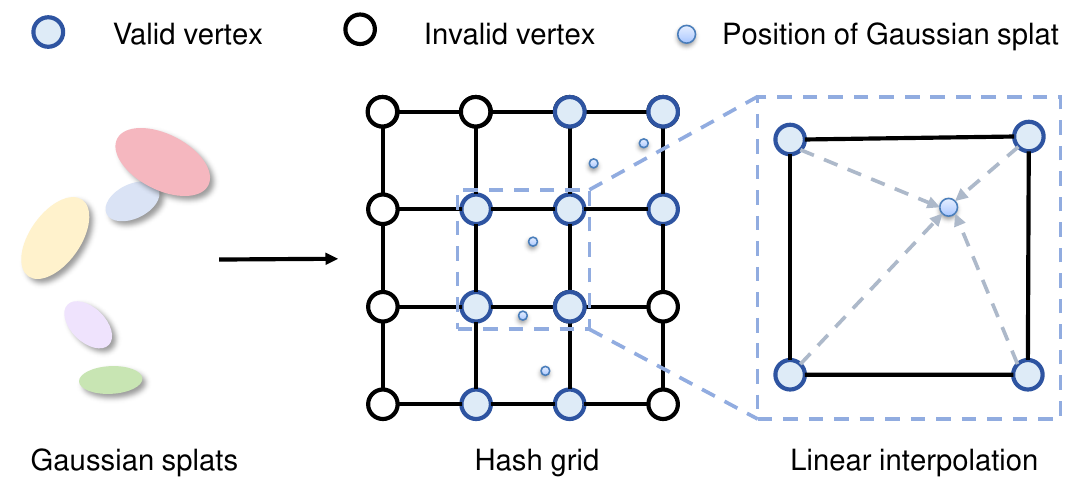}
\caption{\textbf{The valid and invalid vertices in hash grid.} The hash grid leverages the features of vertices surrounding each Gaussian splat to compute the feature at that location via linear interpolation. Due to the non-uniform distribution of Gaussian splats, a large number of vertices in the grid do not participate in computation, and their corresponding features are thus designated as invalid.}
\label{fig:preminarily}
\end{figure}

Gaussian splatting is of great importance for applications such as virtual reality, autonomous driving, robotic navigation, and immersive telepresence, for which its efficient representation and compression are essential for practical deployment. 
Inspired by its successful application in earlier radiance field representations, hash grids~\cite{muller2022instant} have been widely adopted to learn a neural field that captures implicit spatial relationships among Gaussian splats. For example, they serve as structural contextual information to assist entropy modeling for compression~\cite{chen2024hac,chen2025hac++}, or represent a residual field for inter-frame dynamic modeling~\cite{sun20243dgstream,tang2025compressing,wuswift4d}. Typically, these methods take a fixed position set of the targeted Gaussian splats as input to the hash grid at the inference stage. The grid indexes features from vertices surrounding each targeted position, and linear interpolation is then applied to predict the implicit feature at that position, thereby capturing its spatial characteristics.
However, the non-uniform spatial distribution of discrete Gaussian splats results in extensive sparse regions within the 3D space. Consequently, as illustrated in Fig.~\ref{fig:preminarily}, a significant portion of the vertex features in the grid remain invalid, as they never participate in inference. From a compression and transmission perspective, encoding these invalid features introduces unnecessary bitrate overhead, which degrades overall rate-distortion performance.
   
To address the aforementioned issue, this work introduces hash grid pruning, which removes invalid features and encodes only the valid ones, thereby reducing the bitrate. Specifically, we first determine the valid vertex positions required for computation based on the input coordinates. These valid positions are then indexed via a hash function~\cite{muller2022instant} to retrieve their corresponding valid features, while the remaining invalid features are pruned. Since the invalid features contribute nothing to the inference, their removal does not affect the model’s reconstruction capability while solely reducing the bitrate, thus improving the rate‑distortion performance.

\begin{figure}[t]
\centering
\includegraphics[width=0.99\linewidth]{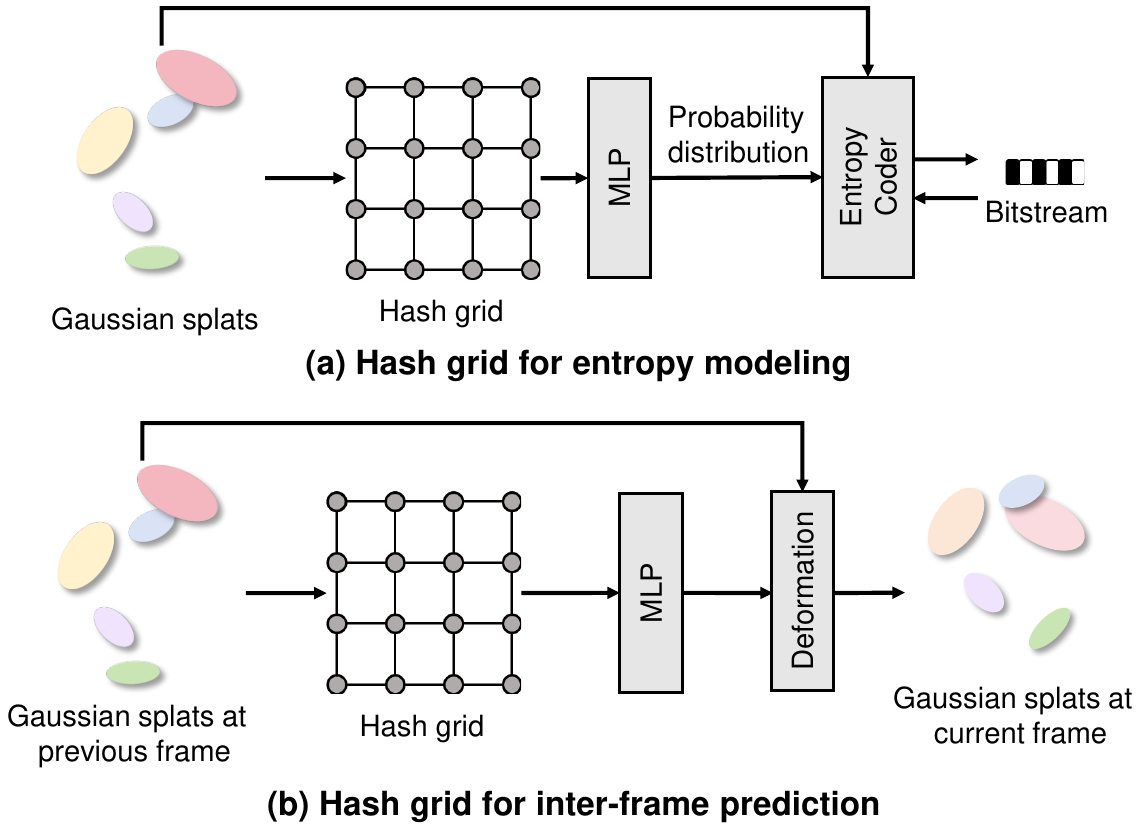}
\caption{\textbf{Hash grid adoption in i3DV.} (a) depicts the use of a hash grid to assist entropy modeling for static Gaussian splat compression. (b) illustrates the employment of a hash grid to represent an implicit residual field for dynamic Gaussian splat compression.}
\label{fig:hashgridini3dv}
\end{figure}

This work is also a technical proposal to the Virtual Reality Union (VRU) subgroup of the Audio Video coding Standard Workgroup of China (AVS). 
All experiments are conducted using the group's integrated test model, the Intelligent 3D Volumetric Video Coding Platform (i3DV)~\cite{AVS-i3DV}. Within this codec, the hash grid is utilized in two key aspects as illustrated in Fig~\ref{fig:hashgridini3dv}. For static Gaussian splat compression, it serves as contextual information to assist entropy modeling, similar to HAC~\cite{chen2024hac}. 
For streamable dynamic Gaussian splat compression, it helps construct an implicit residual field for inter-frame prediction, analogous to iFVC~\cite{tang2025compressing}. We apply the proposed pruning strategy to both processes and evaluate performance under the Common Test Conditions (CTC)~\cite{AVS-VRU-CTC} defined by the committee. Experimental results show that our method achieves up to 15\% and an average of 8\% bitrate saving across all test sequences compared to the baseline.

\section{Method}

To prune redundant features, the primary objective is to distinguish between valid and invalid features based on the known input positions of the Gaussian splats. For simplicity, we illustrate the process using a two-dimensional example, which can be readily extended to higher-dimensional scenarios.

Given a fixed set of input positions $X = (x, y)$ and the resolution $R$ of the hash grid, the scaled position $X_R = (x_R, y_R)$ at resolution $R$ is calculated as:
\begin{equation}
    X_R = \frac{X - X_{\min}}{X_{\max} - X_{\min}} \cdot R,
\end{equation}
where $X_{\min}$ and $X_{\max}$ denote the minimum and maximum values of $X$, respectively.
Since the hash grid employs linear interpolation to predict the implicit residual at $X_R$ using the features stored in the grid, we calculate the set of neighboring vertices $\{X_V\}_{\text{Neighbor}}$ (four vertices in 2D) as follows:
\begin{equation}
    \{X_V\}_{\text{Neighbor}} =
    \left\{
        \begin{array}{@{}c@{}}
            (\lfloor x_R \rfloor + 0, \lfloor y_R \rfloor + 0), \\
            (\lfloor x_R \rfloor + 1, \lfloor y_R \rfloor + 0), \\
            (\lfloor x_R \rfloor + 0, \lfloor y_R \rfloor + 1), \\
            (\lfloor x_R \rfloor + 1, \lfloor y_R \rfloor + 1)
        \end{array}
    \right\}.
\end{equation}
Following the standard indexing scheme of the hash grid~\cite{muller2022instant}, a hash function is applied to map $\{X_V\}_{\text{Neighbor}}$ to the corresponding feature indices. The hash function $h(\cdot)$ is formulated as:
\begin{equation}
    h(X_V) = \big( (x_V \pi_x) \oplus (y_V \pi_y) \big) \bmod T,
\end{equation}
where $\pi_x$ and $\pi_y$ are large prime numbers for each respective dimension, $\oplus$ denotes the bit-wise XOR operation, and $T$ denotes the size of the hash table. The valid features are then selected based on these indices, while the remaining invalid ones are pruned.

\begin{figure}[t]
\centering
\includegraphics[width=0.98\linewidth]{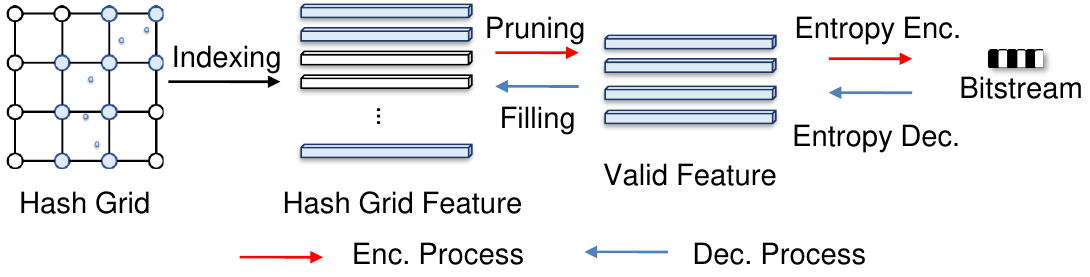}
\caption{\textbf{Combining pruning strategy with compression.} Based on the input positions, invalid features are identified and pruned, leaving only the valid features to be encoded and decoded.}
\label{fig:method}
\end{figure}

As illustrated in Fig.~\ref{fig:method}, after pruning the invalid features, we apply entropy encoding solely to the valid ones to reduce the bitstream size. On the decoder side, the same set of valid feature indices can be calculated based on the decoded positions, allowing the decoded features to be correctly filled into the hash table. Since the invalid features contribute nothing to the inference, their corresponding entries can be assigned arbitrary values without affecting performance. This design reduces the bitrate of the encoded hash grid while preserving reconstruction quality, thereby improving the overall rate-distortion performance.

\begin{table}[]
\centering
\caption{\textbf{Rate reduction after adopting pruning strategy.} Rate is calculated in MB per frame.}
\begin{tabular}{@{}cccc@{}}
\toprule
 & i3DV-3.0-LD & Ours & Reduction Ratio \\ \midrule
\multicolumn{4}{c}{\textit{DanceDunhuang}} \\ \midrule
Rate1 & 0.154 & 0.14 & 10.00\% \\
Rate2 & 0.192 & 0.17 & 12.90\% \\
Rate3 & 0.247 & 0.212 & 17.00\% \\
Rate4 & 0.33 & 0.27 & 22.30\% \\
Average & 0.231 & 0.198 & 15.50\% \\ \midrule
\multicolumn{4}{c}{\textit{ShowGroups}} \\ \midrule
Rate1 & 0.055 & 0.054 & 2.10\% \\
Rate2 & 0.058 & 0.056 & 3.30\% \\
Rate3 & 0.067 & 0.064 & 5.00\% \\
Rate4 & 0.082 & 0.076 & 7.90\% \\
Average & 0.066 & 0.062 & 4.60\% \\ \midrule
\multicolumn{4}{c}{\textit{VRUgz}} \\ \midrule
Rate1 & 0.072 & 0.07 & 2.70\% \\
Rate2 & 0.087 & 0.083 & 4.00\% \\
Rate3 & 0.108 & 0.102 & 5.30\% \\
Rate4 & 0.151 & 0.141 & 6.90\% \\
Average & 0.104 & 0.099 & 4.80\% \\ \midrule
\multicolumn{4}{c}{\textit{VRUdg4}} \\ \midrule
Rate1 & 0.098 & 0.093 & 5.70\% \\
Rate2 & 0.099 & 0.094 & 5.20\% \\
Rate3 & 0.124 & 0.115 & 8.10\% \\
Rate4 & 0.15 & 0.135 & 10.60\% \\
Average & 0.118 & 0.109 & -7.40\% \\ \midrule
\multicolumn{2}{c}{Average reduction across all sequences} & \multicolumn{2}{c}{-8.08\%} \\ \bottomrule
\end{tabular}
\label{tab:ratereduction}
\end{table}

\begin{figure*}[t]
	\centering
	\includegraphics[width=2.9in]{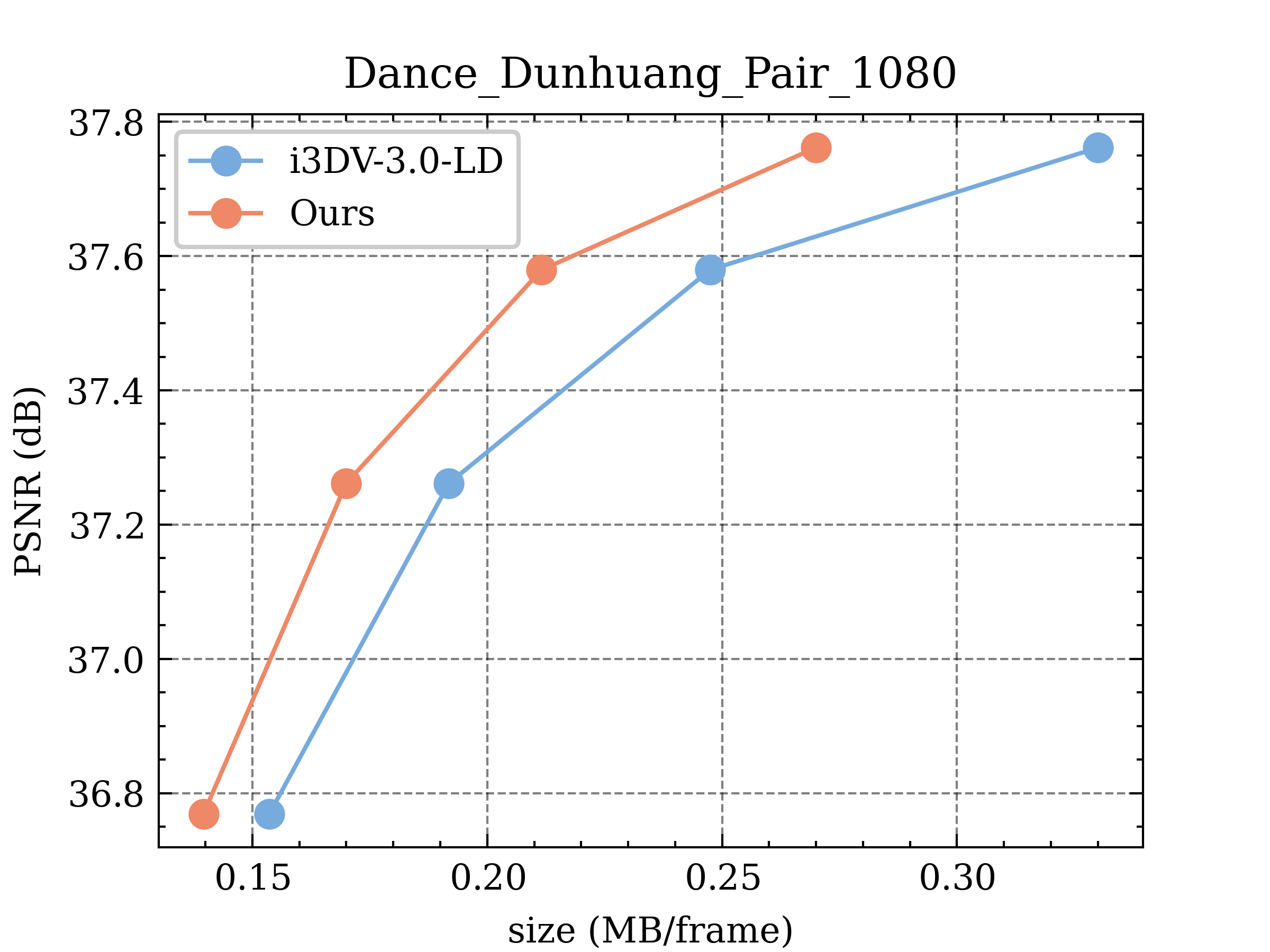}
	\includegraphics[width=2.9in]{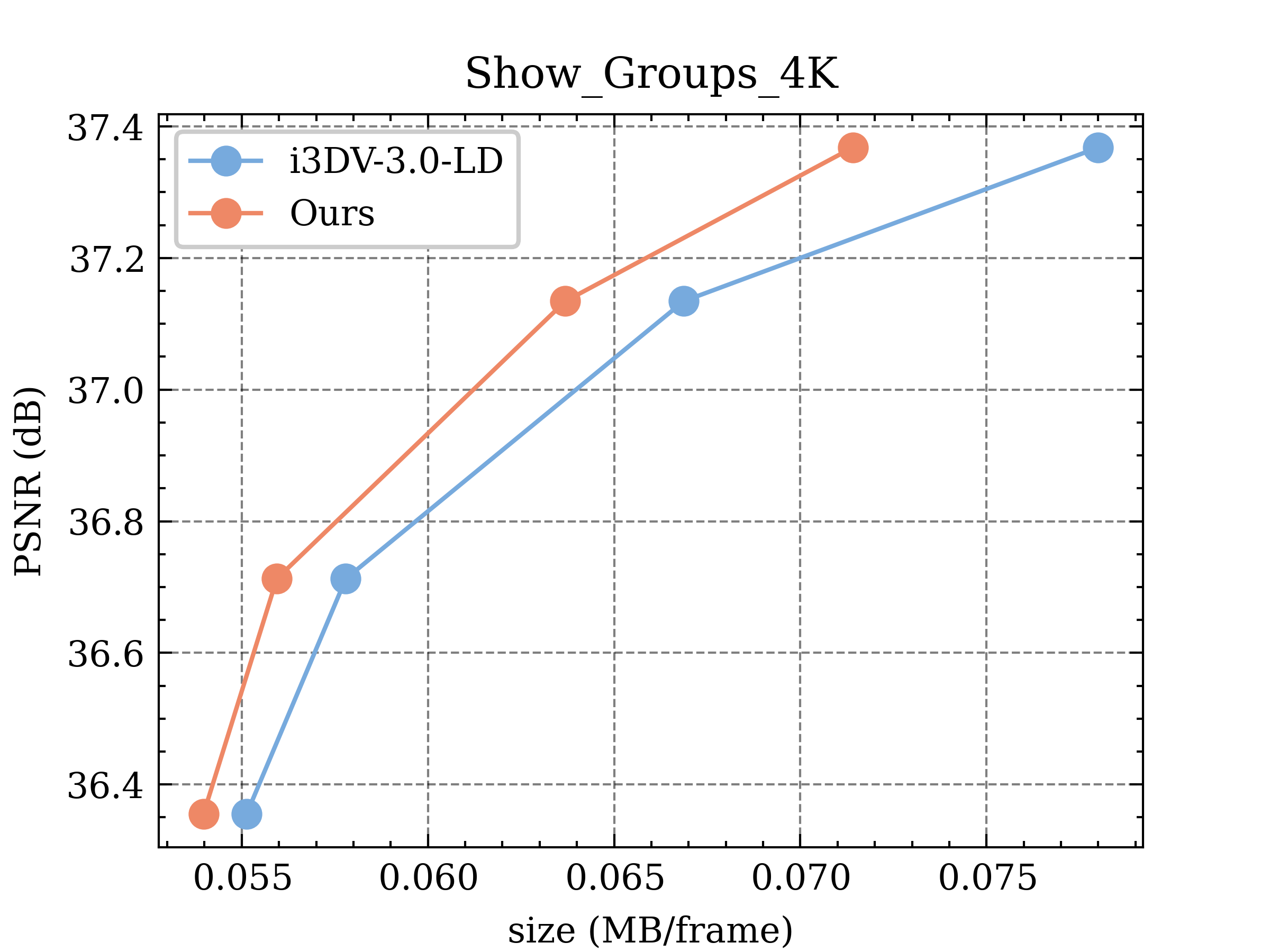}
    \includegraphics[width=2.9in]{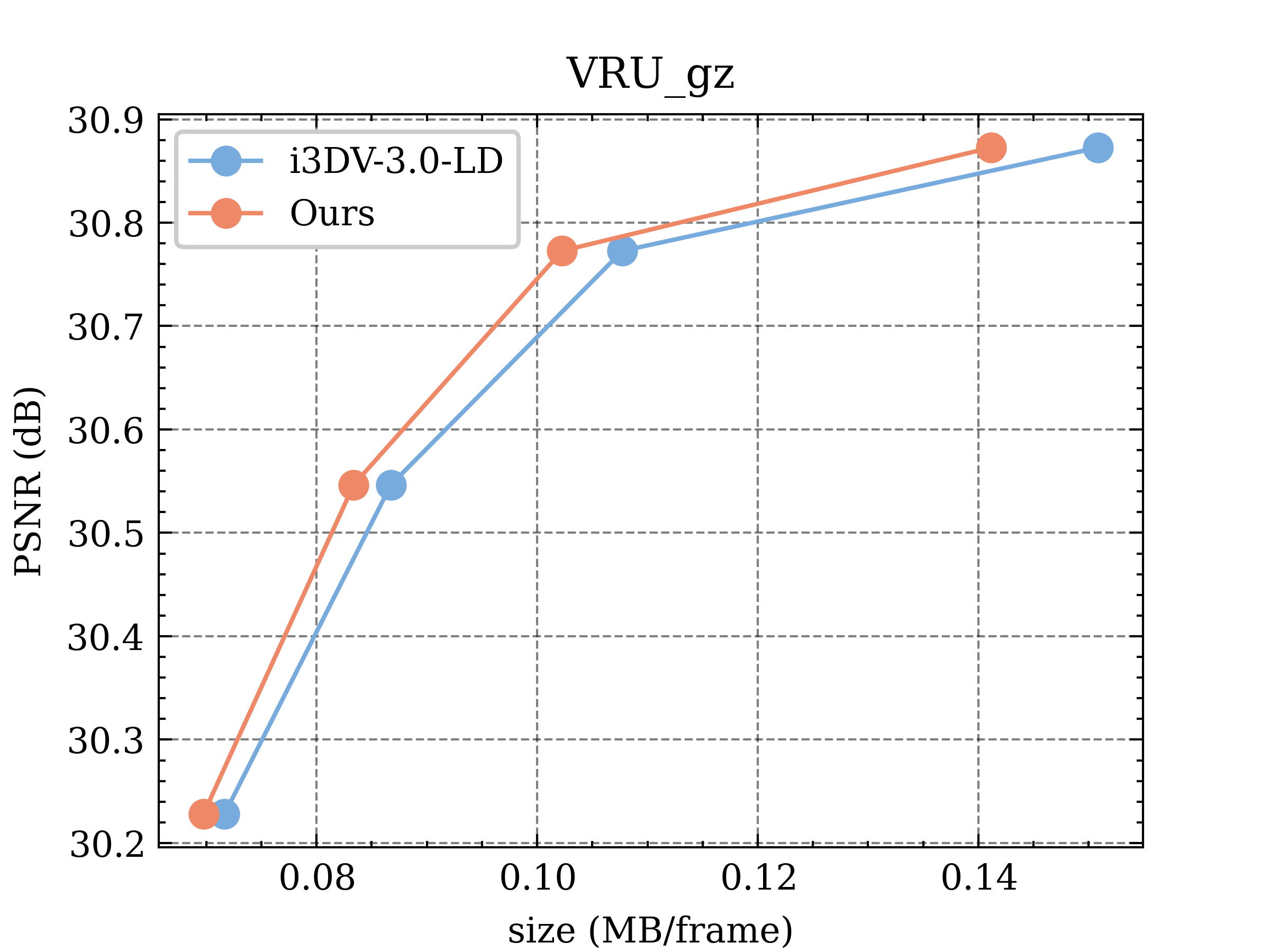}
	\includegraphics[width=2.9in]{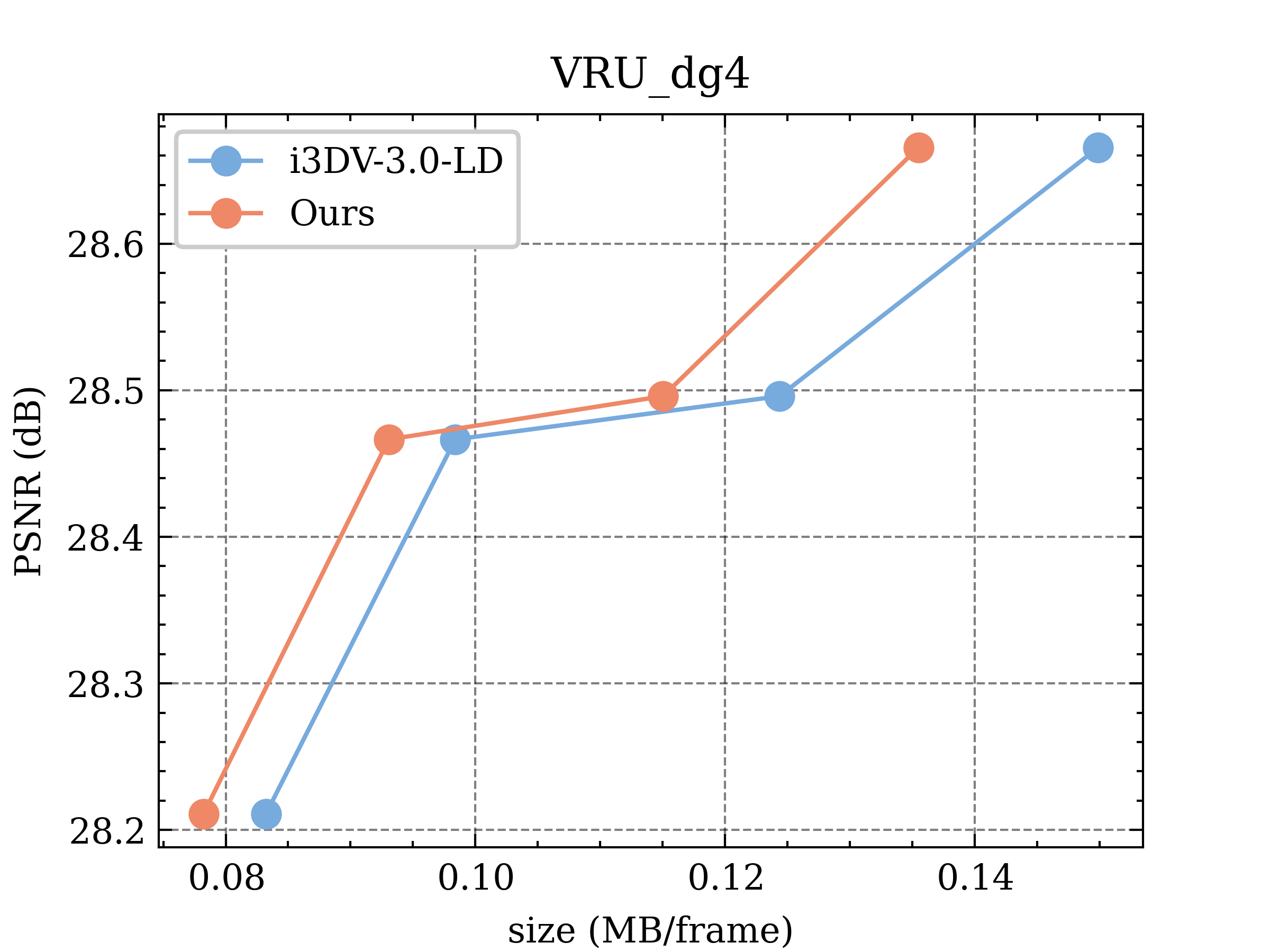}
	\caption{{\bf Rate-distortion Comparison.} Rate-Distortion (R-D) curves of different methods. 250 frames are evaluated across a wide range of size following the CTC~\cite{AVS-VRU-CTC}.}
\label{fig:rdcurves}
\end{figure*}

\section{Experiments}

\subsection{Experimental Conditions}
Experiments are conducted under the CTC defined by the AVS-VRU committee, using four standard multi‑view sequences: the 1080p sequence \textit{DanceDunhuang}, the 4K sequence \textit{ShowGroup}, and two basketball sequences with complex motion details, \textit{VRUgz} and \textit{VRUdg4}.

We implement the proposed pruning strategy on the integrated i3DV~\cite{AVS-i3DV} test model (version 3.0) in low‑delay (LD) mode for both \textit{intra} and \textit{inter} coding scenarios. The training strategy remains identical to that of i3DV-3.0, as pruning is applied as a post‑training procedure.
For rate–distortion evaluation, PSNR (in dB) is used as the distortion metric, and the total file size per frame (in MB) serves as the rate metric.

\subsection{Performance Evaluation}
As shown in Table~\ref{tab:ratereduction}, the proposed hash‑grid feature pruning strategy achieves an average bitrate reduction of 8.08\% across all test sequences, with the highest reduction of 15.5\% attained on the \textit{DanceDunhuang} sequence. 
Furthermore, the performance gain from pruning is more pronounced at higher bitrates. This trend indicates that the redundancy present in the hash grid features is greater under high-bitrate conditions.
Meanwhile, as shown in Fig.~\ref{fig:rdcurves}, the PSNR after pruning remains identical to that of the baseline across all operating points. This result confirms that the proposed pruning strategy does not affect reconstruction performance, thereby preserving the original quality.

\section{Conclusion}

 This technical report introduces a hash‑grid pruning strategy designed as a post‑training step for hash‑grid‑based Gaussian splatting compression. The process begins by distinguishing between valid and invalid grid features according to the input positional set. For storage and transmission, invalid features are pruned, and entropy coding is applied only to the valid ones. Under the CTC of AVS‑VRU, the proposed strategy achieves an average bitrate reduction of 8\% across all test sequences while preserving reconstruction quality.

\newpage
\bibliographystyle{IEEEbib}
\bibliography{refs}

\end{document}